\documentclass{article}

\usepackage{arxiv}

\usepackage[utf8]{inputenc} % allow utf-8 input
\usepackage[T1]{fontenc}    % use 8-bit T1 fonts
\usepackage{hyperref}       % hyperlinks
\usepackage{url}            % simple URL typesetting
\usepackage{booktabs}       % professional-quality tables
\usepackage{amsfonts}       % blackboard math symbols
\usepackage{nicefrac}       % compact symbols for 1/2, etc.
\usepackage{microtype}      % microtypography
\usepackage{lipsum}
\usepackage{amsthm}
\usepackage{graphicx}
\usepackage{acronym}
\usepackage{psfrag}
\usepackage{multirow}
\usepackage{subfigure}
\usepackage{mathtools}
\title{General Supervision via Probabilistic Transformations}

\author{
  Santiago Mazuelas \\
  BCAM-Basque Center of Applied Mathematics\\
  Bilbao, Spain\\
  \texttt{smazuelas@bcamath.org} \\
   \And
 Aritz P\'{e}rez\\
  BCAM-Basque Center of Applied Mathematics\\
  Bilbao, Spain\\
  \texttt{aperez@bcamath.org} \\
}
\date{}
\begin{document}
\maketitle

\theoremstyle{definition}
\newtheorem{definition}{Definition}
\newtheorem{proposition}{Proposition}
\newtheorem{theorem}{Theorem}
\newtheorem{corollary}{Corollary}
\newtheorem{lemma}{Lemma}
\newtheorem{remark}{Remark}
\newcommand\V[1]  { \mathbf{#1} }
\newcommand\B[1]  { \boldsymbol{#1} }
\newcommand\rv[1] {\mathrm{#1}}
\newcommand\set[1] {\mathcal{#1}}
\newcommand\tri { {{\triangle}} }
\newcommand\dia { \Diamond }
\newcommand{\ar}[1]{ xtcolor{sarandonga}{#1}}
\newcommand{\arc}[1]{ xtcolor{sarandonga}{[#1]}}
\newcommand{\arf}[1]{\footnote{ xtcolor{sarandonga}{#1}}}
\acrodef{RV}{random variable}
\acrodef{ERM}{empirical risk minimization}
\acrodef{RRM}{robust risk minimization}
\acrodef{GRRM}{generalized \ac{RRM}}
\acrodef{TES}{Test stage}
\acrodef{TRS}{Training stage}
\acrodef{RB}{Representation based}
\acrodef{SVM}{support vector machine}

%\author{Aritz Perez\\
%Basque Center for Applied Mathematics\\
%2275 East Bayshore Road, Suite 160\\
%Palo Alto, California 94303\\
%}
\begin{abstract}
Different types of training data have led to numerous schemes for supervised classification. Current learning techniques are tailored to one specific scheme and cannot handle general ensembles of training data. This paper presents a unifying framework for supervised classification with general ensembles of training data, and proposes the learning methodology of generalized robust risk minimization (GRRM). The paper shows how current and novel supervision schemes can be addressed under the proposed framework by representing the relationship between examples at test and training via probabilistic transformations. The results show that GRRM can handle different types of training data in a unified manner, and enable new supervision schemes that aggregate general ensembles of training data.
\end{abstract}

\section{Introduction}
Supervised classification uses training data to choose a classification rule with small expected loss over test variables (features and label). Since the probability distribution of test variables is unknown, expected losses are evaluated with respect to a surrogate probability distribution obtained from training data. Approaches based on \ac{ERM} use the empirical distribution of training samples \cite{Vap:98,EvgPonPog:00} while approaches based on \ac{RRM} use a distribution with maximum entropy near the empirical distribution \cite{FarTse:16,ShaKuhMoh:17,AsiXinBeh:15}. 

%In standard supervision, training examples follow the same distribution as test examples, while numerous non-standard supervision schemes have been proposed to exploit more general types of training data.  Current non-standard schemes consider: i) labels at training that are less precise than those at test (e.g., noisy labels \cite{NatDhiRavTew:13,LiuDac:16}, multiple labels \cite{JinGha:03}, and weak multi-labels \cite{SunZhaZho:10}); ii) features at training that are more informative than those at test (e.g., privileged information \cite{PecVap:10} and \ac{TES} corrupted features \cite{DekShaXia:10}); iii) features at training that are less informative than those at test (e.g., \ac{TRS} corrupted features \cite{ShiBhaSmo:06}); iv) examples at training that are from a different domain (e.g., \ac{RB} domain adaptation \cite{ChenWeinBli:11} and covariate shift \cite{SugKaw:12}); v) examples at training with missing components (e.g., semi-supervision \cite{ChaZie:05,QiTiaShi:12} and \ac{TRS} missing features \cite{ShiBhaSmo:06}); and vi) examples at training with multiple qualities and types (e.g., variable quality data \cite{CraKeaWor:06,Roo:15} and domain adaptation with multiple sources \cite{ManMohRos:09}). 
In standard supervision, training examples follow the same distribution as test examples, while numerous non-standard supervision schemes have been proposed to exploit more general types of training data.  Current non-standard schemes consider: i) labels at training that are less precise than those at test; ii) features at training that are more informative than those at test; iii) features at training that are less informative than those at test; iv) examples at training that are from a different domain; v) examples at training with missing components; and vi) examples at training with multiple qualities and domains. Those schemes have been developed under different paradigms such as weak supervision, semi-supervision, privileged information, and domain adaptation (see specific current approaches and associated references in Sections~\ref{Sec_gen_sup} and \ref{Sec_gen_het}). 
%Current heterogenous supervision schemes use: examples at training that follow the same distribution as prediction examples together with examples that miss labels (semi-supervision \cite{ChaZie:05,QiTiaShi:12}) or feature components (\ac{TS} missing features \cite{ShiBhaSmo:06}); examples at training of multiple types such as labels with different noise intensities (variable quality data \cite{CraKeaWor:06,Roo:15}) or features from different domains (domain adaptation with multiple sources \cite{ManMohRos:09}).

The diverse range of supervision schemes described above have shown to be extremely useful in practice. Schemes that use training examples from different domains or less precise than test examples can reduce training costs, while those that use training examples more precise than test examples can increase classification accuracies. Current techniques are tailored to one specific supervision scheme and there is a lack of a common methodology for supervised classification with general training data. As a consequence, it is currently not possible to adequately deal with cost/accuracy trade-offs and to seamlessly develop versatile algorithms. For instance, existing techniques can only handle scenarios with training data in accordance with one of the specific cases described above, and cannot exploit general ensembles of training data with assorted types and qualities. This paper presents a unifying framework for supervised classification with general ensembles of training data, and proposes the learning methodology of \ac{GRRM}. Such framework is enabled by representing the relationship between examples at test and training stages via probabilistic transformations. The paper shows how current and novel supervision schemes can be addressed under the proposed framework. In particular, we show that \ac{GRRM} can enable learning algorithms that aggregate general ensembles of training data with different types.
\section{Preliminaries}\label{sec_standard}
This section provides an overview of the supervised classification problem, recalls the notion of probabilistic transformation, and describes notations used in the rest of the paper. In particular, in the following, upright upper case letters denote \acp{RV}; calligraphic upper case letters denote sets; $\mathbb{I}\{\cdot\}$ denotes the indicator function;  $\mathbb{E}_{a\sim P}\{f(a)\}$ or just $\mathbb{E}_{P}\{f(a)\}$ denotes the expectation of function $f$ over instantiations $a$ that follow probability distribution $P$; and $I$ denotes an identity transformation. 
\subsection{Supervised classification} 
A problem of supervised classification can be described by four objects $(\rv{Z},\rv{D},\set{H},L)$ representing  variables at test, training data, classification rules, and miss-classification losses. Specifically, $\rv{Z}=(\rv{X},\rv{Y})$ is an \ac{RV} representing examples at test, $\rv{X}$ is called feature or attribute, and $\rv{Y}$ has finite support and is called label or class. $\rv{D}$ is an \ac{RV} describing training data formed by the concatenation of training samples. For instance, in standard supervision each instantiation of $\rv{D}$ is $d=(z^{(1)},z^{(2)},\ldots,z^{(n)})$ where $z^{(i)}$ for $i=1,2,\ldots,n$ are independent instantiations of $\rv{Z}$. 
The classification rules $\set{H}$ are mappings from features to labels, i.e., $h\in\set{H}$, $h:\set{X}\to\set{Y}  $. Finally,
$L$ is a function $L:\set{Y}\times \set{Y}\to\mathbb{R}$,  where $L(\hat{y},y)$ quantifies the loss of predicting the label $y$ by label $\hat{y}$, e.g., $L(\hat{y},y)=\mathbb{I}\{y\neq\hat{y}\}$. 

The goal of a learning algorithm for classification is to determine a rule $h\in\set{H}$ with small expected loss (risk) under the probability distribution of $\rv{Z}$, $P$, that is, to solve the optimization problem 
%\begin{align}\label{min_risk}h^*\in\arg\min_{h\in\set{H}}\mathbb{E}_p\{\ell_h(x)\}\end{align} 
\begin{align}\label{min_risk}\min_{h\in\set{H}}\mathbb{E}_P\big\{L(h(x),y)\big\}.\end{align} 
%with $\ell_h$ the loss function for classification rule $h$, given by $\ell_h(x)=C(z,h(y))$ for $x=(y,z)$. 
Training data aids the learning problem in that it provides information regarding the probability distribution $P$.
% Such problem can be seen as a zero-sum game between learner and nature in which, learner selects a classification rule from $\set{H}$, after which nature reveals the value of $x\in\set{X}$, and learner suffer a loss $\ell_h(x)$ [cite game theory, maximum entropy...] Nature strategy can be represented by the probability distribution of $\rv{X}$, and the undesirability to learner of any classification rule $h\in\set{H}$ is assessed by means of its expected loss [cite game theory, maximum entropy...] $\mathbb{E}_p\{\ell_h(x)\}$ with $p$ the probability distribution of $\rv{X}$. 
%After observing training data $y\in\set{Y}$ during supervision,  given $y$, and the undesirability to learner of any classification rule $z\in\set{Z}$ is assessed by means of its expected loss [cite game theory, maximum entropy...]
%\begin{align}\label{exp_loss}\mathbb{E}_p\{\ell_z(x)\}\end{align}
%with $p$ the probability distribution of $\rv{X}$ given $y$. 

Supervised learning based on \ac{ERM} corresponds to solving \eqref{min_risk} using the empirical distribution $P_e$ of the training data $d$ as surrogate for $P$. The main drawback of \ac{ERM} approach is over-fitting that is addressed using regularization methods. Most techniques for regularization are based on structural \ac{ERM} that considers subsets of classification rules with reduced complexity \cite{Vap:98,EvgPonPog:00}. Other complementary regularization techniques are based on \ac{RRM} that considers uncertainty (ambiguity) sets $\set{U}$ of probability distributions \cite{FarTse:16,ShaKuhMoh:17,AsiXinBeh:15}. Specifically, the classification rule in such techniques is obtained by minimizing the maximum expected loss over the uncertainty set, i.e., solving
%\begin{align}&\label{reg_ent}\min_{h\in \set{H}}\max_{q\in\set{U}}\mathbb{E}_{q}\{\ell_{h}(x)\}\end{align}
\begin{align}&\label{reg_ent}\min_{h\in \set{H}}\max_{Q\in\set{U}}\mathbb{E}_{Q}\big\{L(h(x),y)\big\}\,.\end{align}%\\\mbox{with }%&
%\label{un_set}\set{U}=\{p\in\tri(\set{X}):\  \psi(p,p_d)\leq\varepsilon\}\end{align} for a function
The uncertainty set $\set{U}$ is formed by distributions close to the empirical distribution, where the closeness between distributions in $\set{Z}$ can be quantified by a discrepancy function $\psi$, hence
$$\set{U}=\left\{Q\in\tri(\set{Z}):\  \psi(Q,P_e)<\varepsilon\right\}$$
with $\tri(\set{Z})$ the set of probability distributions supported in $\set{Z}$. For instance, the uncertainty sets used in \cite{ShaKuhMoh:17} correspond to consider as
$\psi(Q_1,Q_2)$ the Wasserstein (transportation) distance between $Q_1$ and $Q_2$, while those used in \cite{FarTse:16} correspond to 
\begin{align}\label{tse}\psi(Q_1,Q_2)=\|\mathbb{E}_{Q_1}\{t(z)\}- \mathbb{E}_{Q_2}\{t(z)\}\|\end{align}
for $Q_1$ and $Q_2$ distributions with the same marginal over $\set{X}$, and $t(\cdot)$ a statistic over $\set{Z}$.

For each distribution $Q\in\tri(\set{Z})$, the minimum expected loss defines an entropy function as $H(Q)=\min_{h\in\set{H}}\mathbb{E}_Q\big\{L(h(x),y)\big\}$ \cite{GruDaw:04}. For instance, if $L(\hat{y},y)=\mathbb{I}(y\neq\hat{y})$ and $\set{H}$ contains the posterior Bayes rule, the entropy is known as 0-1 entropy and is given by 
\begin{align}\label{01ent}H(Q)&=\mathbb{E}_{Q}\left\{1-\max_{y\in\set{Y}  }Q(y|x)\right\}\nonumber\\
&=1-\int\max_{y\in\set{Y}}Q(x,y)dx\end{align}%=1-\int\max_{y\in\set{Y}}p(x,y)dx\end{align}
where $Q(y|x)$ denotes the conditional distribution of $\rv{Y}$ given $\rv{X}$ for $Q$. 
Under mild regularity conditions \cite{FarTse:16,GruDaw:04}, the minimax solution of \eqref{reg_ent} coincides with its maximin solution. Therefore, \ac{RRM} methods solve \eqref{reg_ent} using as surrogate of $Q$ the distribution $Q^*$ that maximizes the associated entropy near the empirical distribution, i.e.,
\begin{align}\label{RRM}Q^*=\arg\min_{Q}\psi(Q,P_e)-\lambda H(q)\end{align}
for a regularization parameter $\lambda$. 
Both \ac{ERM} and \ac{RRM} strategies are often equivalent \cite{BerCop:17}. However, the empirical distribution of non-standard training data is often not adequate to assess the uncertainty about test variables (see Section~\ref{sec_current} below), and in this paper we extend the \ac{RRM} approach for non-standard supervision. 
\subsection{Probability distributions and probabilistic transformations}
Probabilistic transformations, also known as Markov transitions or just transitions \cite{ChaKim:94,RooWill:18}, are a generalization of the concept of deterministic transformation and allow to represent random and uncertain processes. In the following, for each support set $\set{V}$, a probability distribution $Q\in\tri(\set{V})$ is given by a function on $\set{V}$, e.g., density function or probability mass function.\footnote{We consider \acp{RV} with probability measures dominated by a base measure. More general scenarios can be analogously treated by requiring certain measure-theoretic regularity conditions such as Borel probability measures and Polish spaces, see for instance \cite{ChaKim:94,GruDaw:04}.}
\begin{definition}
%A probabilistic transformation from a set $\set{V}$ to a set $\set{W}$ is a function $T$ that maps each $v\in\set{V}$ to a probability distribution $T(v)\in\tri(\set{W})$. Such transformations can be equivalently seen as linear maps from $\tri(\set{V})$ to $\tri(\set{W})$, denoted $\tri(\set{V}, \set{W})$, by defining $T(p)\coloneqq\mathbb{E}_{v\sim p} T(v)\in\tri(\set{W})$ for each $p\in\tri(\set{V})$.
A probabilistic transformation is a linear map that transforms probability distributions into probability distributions. For support sets $\set{V}$ and $\set{W}$, we denote by $\tri(\set{V}, \set{W})$ the set of probabilistic transformations $T$ with $T(Q)\in\tri(\set{W})$ for $Q\in\tri(\set{V})$.
%from $\tri(\set{V})$ to $\tri(\set{W})$a set $\set{V}$ to a set $\set{W}$ are linear maps from $\tri(\set{V})$ to $\tri(\set{W})$
\end{definition}

%element $v\in\set{V}$ is a uniform probability distribution with support $f(v)$. % For instance, the function $\Pi_x(x,y)=x$ corresponds to the probabilistic transformation $T_{\Pi_x}$ that maps joint distributions in $(x,y)$ into their marginals in $x$.  
If $\set{V}$ and $\set{W}$ have $n$ and $m$ elements, respectively,  a probabilistic transformation in $\tri(\set{V},\set{W})$ is given by a $n\times m$ row-stochastic Markov transition matrix $K$; then $T(Q)=R$ given by $R(w)=\sum_{v\in\set{V}} K(v,w)Q(v)$, with $K(v,w)$ the matrix component in row $v\in\set{V}$ and column $w\in\set{W}$. Analogously, for infinite sets, a probabilistic transformation in $\tri(\set{V},\set{W})$ is given by a function $K(v,w)$ called Markov transition kernel, then $T(Q)=R$ given by $R(w)=\int_\set{V}K(v,w)Q(v)dv$.
Simple examples of probabilistic transformations are deterministic and set-valued functions $f:\set{V}\to\set{W}$ in which the image of a distribution supported in a single point $v$ is a uniform probability distribution with support $f(v)$. In addition, the conditional distribution of an \ac{RV} $\rv{W}$ conditioned on an \ac{RV} $\rv{V}$ provides a probabilistic transformation denoted $T_{\rv{W}|\rv{V}}$ that maps the probability distribution of $\rv{V}$ to that of $\rv{W}$.
\begin{figure*}
	\centering
%	\psfrag{Random variable values}[c][][1]{\hspace{-1mm}RV values}
\psfrag{a}[l][][1]{\hspace{-3.5mm}$\tri(\set{Z})$}
\psfrag{b}[l][][1]{\hspace{-3.5mm}$\tri(\tilde{\set{Z}})$}
\psfrag{c}[l][][1]{\hspace{-3.5mm}$\tri(\set{B})$}
\psfrag{t1}[l][][1]{\hspace{0mm}$T$}
\psfrag{t2}[l][][1]{\hspace{0mm}$\tilde{T}$}
\psfrag{p}[l][][0.8]{\hspace{6.5mm}$P$}
\psfrag{u1}[l][][0.8]{\hspace{-5mm}$\set{U}$}
\psfrag{u2}[l][][0.8]{\hspace{-3.5mm}$\tilde{\set{U}}$}
\psfrag{q}[l][][0.8]{\hspace{5mm}$\tilde{P}$}
\psfrag{pd}[l][][0.8]{\hspace{7mm}$\tilde{P}_e$}
\psfrag{r}[l][][0.8]{\hspace{-7mm}$T(P)$}
\psfrag{s}[l][][0.8]{\hspace{-6.5mm}$\tilde{T}(\tilde{P})$}
	%	\psfrag{Delta}[l][][0.7]{\hspace{-2mm}$\tri_n$}
	\includegraphics[width=0.95\textwidth] {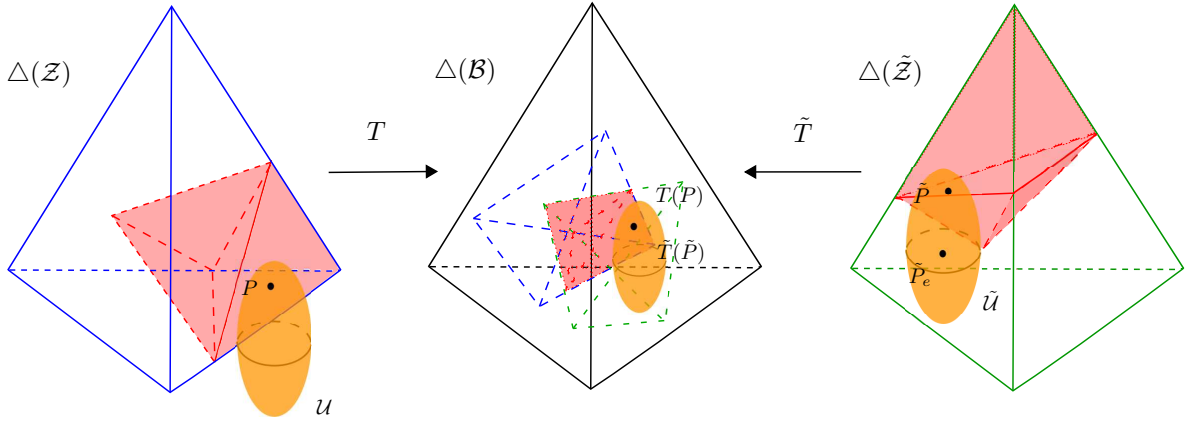}
\caption{The relationship between test and training variables imposes structural constraints for feasible distributions (red polyhedra), and enables to use information from training data as uncertainty sets (orange ellipsoids).}\label{fig1} %Specifically, feasible distributions
% $p\in\tri(\set{X})$ satisfy $T(p)=\tilde{T}(\tilde{p})$ for a $\tilde{p}\in\tri(\tilde{\set{X}})$, and a uncertainty set $\tilde{\set{U}}$ given by the empirical distribution of data $\tilde{p}_d$ determines a uncertainty set in $\set{U}$.}\label{fig1}	
\end{figure*}
 %, and the image of a probability distribution $p$ given by a row vector in the $n$-simplex is given by the product $pT$. In what follows, we will use such notation also for the general case and denote $T(p)$ as $pT$ [needed?].

Probabilistic transformations can be composed in series and in parallel. For instance, if $T_1\in\tri(\mathcal{V}_1,\mathcal{W}_1)$ and $T_2\in\tri(\mathcal{V}_2,\mathcal{W}_2)$ are given by Markov transitions kernels $K_1(v_1,w_1)$ and $K_2(v_2,w_2)$, respectively, the parallel composition of $T_1$ and $T_2$ denoted $T_1\otimes T_2\in\tri(\mathcal{V}_1\times\mathcal{V}_2,\mathcal{W}_1\times\mathcal{W}_2)$ is given by the Markov transition kernel $K_1(v_1,w_1)K_2(v_2,w_2)$. %, for any $p\in \tri(\mathcal{V}_1\times\mathcal{V}_2)$, by 
%\begin{align*}T_1\otimes T_2(p)=\mathbb{E}_{p}\{T_1(v_1)\otimes T_2(v_2)\}\in\tri(\mathcal{W}_1\times\mathcal{W}_2)\end{align*}  where $T_1(v_1)\otimes T_2(v_2)$ denotes the joint distribution given by the product of probability distributions $T_1(v_1)$ and $T_2(v_2)$. 
For finite support sets, composition in series and parallel corresponds to matrix multiplication and Kronecker product, respectively.
\section{Supervision with non-standard training data}\label{Sec_gen_sup}
In this section we consider non-standard supervision cases in which examples at test and training are instantiations of two possibly different \acp{RV} $\rv{Z}$ and $\tilde{\rv{Z}}$, that is, training data are $d=(\tilde{z}^{(1)},\tilde{z}^{(2)},\ldots,\tilde{z}^{(n)})$  where $\tilde{z}^{(i)}$ for $i=1,2,\ldots,n$ are independent instantiations of $\tilde{\rv{Z}}$. Several current supervision schemes use non-standard training data such as:
\begin{itemize}%[noitemsep,topsep=0pt]
\item Noisy labels \cite{NatDhiRavTew:13,LiuDac:16}: labels at test and training take the same categorical values, but training labels are affected by errors.
\item Multiple labels \cite{JinGha:03}: labels at test are single categorical values and labels at training are sets of categorical values.  
\item Weak multi-labels \cite{SunZhaZho:10}: labels at test are sets of categorical values and labels at training are partial sets of categorical values.
\item Privileged information \cite{PecVap:10}: features at training stage have more components than those at test.
\item \ac{TES} corrupted features \cite{DekShaXia:10}: features at test are corrupted by noise.
\item \ac{TRS} corrupted features  \cite{ShiBhaSmo:06}: features at training are corrupted by noise.
\item \ac{RB} domain adaptation \cite{ChenWeinBli:11}: examples at test and training belong to different domains that share a common representation.
\item Covariate shift \cite{SugKaw:12}: variables at test and training share the same conditional distribution of labels given features, but features at test and training have different marginal distributions.
\end{itemize}
In the following we present a unifying framework for non-standard supervision, and describe how current and novel schemes can be addressed under such framework.
%\begin{itemize}
%\item the serial composition of $T_1\in\tri(\mathcal{X}_1,\mathcal{X}_2)$ and $T_2\in\tri(\mathcal{X}_2,\mathcal{X}_3)$ denoted $T_1T_2\in\tri(\mathcal{X}_1,\mathcal{X}_3)$ is given, for any $p\in \tri(\set{X}_1)$, by
%\begin{align*}
%p(T_1T_2)=T_2(T_1(p))=\mathbb{E}_{pT_1}T_2(x_2)\in\tri(\mathcal{X}_3)\end{align*} 
%\item the parallel composition of $T_1\in\tri(\mathcal{X}_1,\mathcal{X}_2)$ and $T_2\in\tri(\mathcal{Y}_1,\mathcal{Y}_2)$ denoted $T_1\otimes T_2\in\tri(\mathcal{X}_1\times\mathcal{Y}_1,\mathcal{X}_2\times\mathcal{Y}_2)$ is given, for any $p\in \tri(\mathcal{X}_1\times\mathcal{Y}_1)$, by 
%\begin{align*}p(T_1\otimes T_2)=\mathbb{E}_{p}\{T_1(x_1)\otimes T_2(y_1)\}\in\tri(\mathcal{X}_1\times\mathcal{Y}_1,\mathcal{X}_2\times\mathcal{Y}_2)\end{align*}  where $T_1(x_1)\otimes T_2(y_1)$ denotes the joint distribution given by the product of probability distributions $T_1(x_1)$ and $T_2(y_1)$.  
%\end{itemize}
\subsection{Unifying framework for non-standard supervision}\label{Sec_framework}
%The following shows how supervised classification can be carried out with non-standard training data using the proposed framework based on probabilistic transformations. %In particular, the relationship between variables at prediction and training stages can be represented by means of up to two probabilistic transformations that map the distributions of prediction and training variables into a common distribution in a possibly different space. 
Let $\set{B}$ be a support set, and $T\in\tri(\set{Z},\set{B})$ and $\tilde{T}\in\tri(\tilde{\set{Z}},\set{B})$ be probabilistic transformations such that $T(P)=\tilde{T}(\tilde{P})$ for $P$ and $\tilde{P}$ the distributions of $\rv{Z}  $ and $\tilde{\rv{Z}}$, respectively. $T(P)=\tilde{T}(\tilde{P})$ is the probability distribution of an \ac{RV} $\rv{B}$ that we call \emph{bridge} since it serves to extract information for $\rv{Z}$ from training data in $\tilde{\rv{Z}}$. For instance, if test examples are affected by noisy features and training examples are affected by noisy labels, a variable composed by noisy features and noisy labels can serve as a bridge to extract the information in training data (see third case study in Section~\ref{Sec_exp}). 
%For instance, examples composed by noisy features and noisy labels can serve as bridge for test examples with noisy features and training examples with noisy labels.
Probabilistic transformations $T$ and $\tilde{T}$ represent the relationship between test and training variables, impose structural constrains into the distributions considered, and allow to extract the information in non-standard training data as follows (see Fig.~\ref{fig1}). Feasible distributions $\set{F}  \subset\tri(\set{Z})$ and $\tilde{\set{F}}\subset\tri(\tilde{\set{Z}})$ are
$$\set{F}  =\{Q\in\tri(\set{Z}):\  \exists\, \tilde{Q}\in\tri(\tilde{\set{Z}}),\  T(Q)=\tilde{T}(\tilde{Q})\}  $$
$$\tilde{\set{F}}=\{\tilde{Q}\in\tri(\tilde{\set{Z}}):\  \exists\, Q  \in\tri(\set{Z}),\  \tilde{T}(\tilde{Q})=T(Q)\}$$
Note that feasibility is a necessary condition to be the actual distribution of $\rv{Z}  $ or $\tilde{\rv{Z}}$.  One consequence of the above is that \ac{ERM} approach may be inadequate in these settings since the empirical distribution of training data is often not feasible (see discussion for Equation \eqref{neg} in Section~\ref{sec_current} below). 

The above probabilistic transformations also allow to define uncertainty sets $\set{U}\subset\tri(\set{Z})$ as
$$\set{U}=\{Q\in\tri(\set{Z}):\  \psi(T(Q),\tilde{T}(\tilde{Q}_e))<\varepsilon\}$$
where $\psi$ is a discrepancy function in $\tri(\set{B})$ and  $\tilde{Q}_e$ is the empirical distribution in $\tri(\tilde{\set{Z}})$ of samples $d$. 
Therefore, learning from non-standard training data $d$ can be approached analogously to \ac{RRM}, substituting optimization in \eqref{RRM} by \begin{align}\label{gen_supervision}\min_{Q\in\set{F}} \psi\big(T(Q),\tilde{T}(\tilde{P}_e)\big)-\lambda H(Q)\end{align}
where $\lambda>0$ is a regularization parameter. We call \ac{GRRM} the approach given by using \eqref{gen_supervision} above instead of \eqref{RRM}. Note that it reduces to \ac{RRM} in the case of standard supervision, i.e., $\rv{Z}  =\tilde{\rv{Z}}$, but allows also to use non-standard training data via the structural constraints and uncertainty sets given by the probabilistic transformations $T$ and $\tilde{T}$. 
\begin{table*}
\caption{Current non-standard supervision schemes.\label{table1}}
\centering
\begin{tabular}{llll}
\hline\hline
Supervision scheme					&\hspace{-0.2cm}\begin{tabular}{l}Test $\rv{Z}=(\rv{X},\rv{Y})$\\ vs\\ training $\tilde{\rv{Z}}=(\tilde{\rv{X}},\tilde{\rv{Y}})$\end{tabular}	\hspace{-0.2cm}		&Bridge $\rv{B}$					&\hspace{-0.2cm}\begin{tabular}{l}Prob.\\ transformations\end{tabular}\hspace{-0.2cm}\\\hline\midrule
Noisy labels&\multirow{3}{*}{$\rv{X}=\tilde{\rv{X}}$\begin{tabular}{l}$\tilde{y}$ noisy \\$\tilde{y}$ set, $y\in\tilde{y}$\\$y,\tilde{y}$ sets, $\tilde{y}\subset y$\end{tabular}}\hspace{-0.2cm}&\multirow{3}{*}{$\tilde{\rv{Z}}$}&\multirow{3}{*}{\hspace{-0.2cm}\begin{tabular}{l}$T=I\otimes T_{\tilde{\rv{Y}}|\rv{Y}}$\\$\tilde{T}=I$\end{tabular}}\\
Multiple labels&												&			&										\\
Weak multi-labels&											& 			&										\\\midrule
Privileged information&$\tilde{\rv{X}}=(\rv{X},\rv{X}^{\text{priv}})$ &\multirow{2}{*}{$\rv{Z}$}&\multirow{2}{*}{\hspace{-0.2cm}\begin{tabular}{l}$T=I$\\$\tilde{T}=T_{\rv{X}|\tilde{\rv{X}}}\otimes I$\end{tabular}}\vspace{0.1cm}\\
\ac{TES} corrupted features& $x$ noisy& &\\\midrule
\ac{TRS} corrupted features&$\tilde{x}$ noisy &$\tilde{\rv{X}}$ &\hspace{-0.2cm}\begin{tabular}{l}$T=T_{\tilde{\rv{X}}|\rv{X}}\otimes I$\\$\tilde{T}=I$\end{tabular}\\\midrule
\ac{RB} domain adaptation&$\set{Y}=\tilde{\set{Y}},\  \rv{Y}\neq\tilde{\rv{Y}}$&General&$T=\tilde{T}=T_{\rv{B}|\rv{Z}}$\vspace{0.05cm}\\\hline\hline
\end{tabular}
\end{table*}

The implementation complexity of \ac{GRRM} is also similar to that of \ac{RRM} since their main difference lies on how the uncertainty set $\set{U}$ is defined (by means of $\psi(T(Q),\tilde{T}(\tilde{P}_e))$ instead of $\psi(Q,P_e)$). Therefore, efficient implementations of \ac{GRRM} can be devised similarly as for \ac{RRM}, for instance by exploiting equivalent reformulations based on convex duality \cite{FarTse:16,ShaKuhMoh:17,AsiXinBeh:15}. The determination of transformations $T$ and $\tilde{T}$ in practice requires certain knowledge about the relationship between test and training variables, and possibly to estimate certain parameters similarly to current techniques, e.g., label noise probabilities \cite{NatDhiRavTew:13,LiuDac:16}. This is to be expected since non-standard supervision uses information from training variables that is used for test variables. Note that in most scenarios, such as those described in Tables \ref{table1} and \ref{table-het} below, the knowledge required to determine transformations $T$ and $\tilde{T}$ is quite modest since the same transformations can be used with independence of the actual probability distributions of test and training variables.
\subsection{Different non-standard supervision schemes under the proposed framework}\label{sec_current}
Table~\ref{table1} shows how different current supervision schemes can be addressed under the proposed framework, and how the probabilistic transformations $T$ and $\tilde{T}$ represent the relationship between test and training variables. In certain supervision schemes, such as noisy labels, multiple lables, and \ac{TRS} corrupted features, examples at training stage are less precise than those at test. Then, we can take $\rv{B}=\tilde{\rv{Z}}$ and $T\in\tri(\set{Z},\tilde{\set{Z}})$ the probabilistic transformation corresponding to the conditional distribution of training variables given test variables. In other schemes, such as privileged information and \ac{TES} corrupted features, examples at training stage are more precise than those at test. Then, we can take $\rv{B}=\rv{Z}$ and $\tilde{T}\in\tri(\tilde{\set{Z}},\set{Z})$ the probabilistic transformation corresponding to the conditional distribution of test variables given training variables. Yet in other schemes, such as \ac{RB} domain adaptation, examples at test and training stages are not related by being more or less precise but can be related through a features' representation. Then, we can take $\rv{B}$ as such common representation and $\tilde{T}=T\in\tri(\set{Z},\set{B})$ the probabilistic transformation corresponding to the function mapping features to their representation. 

The proposed framework can offer a common methodology for learning using non-standard training data based on \ac{GRRM} that uses distribution $Q^*$ in \eqref{gen_supervision} as surrogate for $P$ in \eqref{min_risk}. In addition, such framework can bring new insights for the  
design of algorithms for supervised classification. For instance, certain existing approaches for noisy labels \cite{NatDhiRavTew:13,RooWill:18} first transform loss functions in $\set{Z}$ into loss functions in $\tilde{\set{Z}}$ and then use the \ac{ERM} approach in $\tilde{\set{Z}}$. However, the empirical distribution of the training data $\tilde{P}_e$ cannot correspond in this case with a feasible distribution in $\tri(\set{Z})$, because $T(Q)=\tilde{T}(\tilde{P}_e)$ with $\tilde{T}=I$ requires that $Q$ takes both positive and negative values. Specifically, if $\set{Z}=\{-1,+1\}$,
\begin{align}\label{noisy_matrix}T=I\otimes\left[\begin{array}{cc}1-\rho^-&\rho^-\\\rho^+&1-\rho^+\end{array}\right]\end{align}
with $\rho^-$ and $\rho^+$ the probabilities of erroneous labelling in training when the actual label is $-1$ and $+1$, respectively.%\footnote{Equation \eqref{noisy_matrix} assumes that label errors are independent of the features, more general cases can be treated analogously with error probabilities depending on features.} 
Hence, if $T(Q)=\tilde{P}_e$ and $x^{(i)}$ is an instance incorrectly labelled in training as $\tilde{y}=-1$, then \eqref{noisy_matrix} implies that 
\begin{align}\label{neg}
Q(x^{(i)},y=1)&=-\frac{\rho^+}{n(1-\rho^--\rho^+)}%\\
%p(y^{(i)},z=-1)&=-\frac{\rho^-}{n(1-\rho^--\rho^+)} 
\end{align}
that can be significantly smaller than zero for moderate training sizes. This example illustrates that \ac{ERM} can be inadequate for noisy labels, since it determines an optimal classification rule with respect to a measure that is not a probability measure. 
%Equation \eqref{noisy_matrix} assumes that label errors are independent of the features, more general cases can be treated analogously with error probabilities depending on features.}
%\footnote{Cases with label errors dependent on features can be treated analogously with varying error probabilities for different features.}

The presented framework can also enable the development of novel supervision schemes of practical interest. For instance, supervision schemes in which labels at training are more precise than labels at test (e.g., multi-option classification with precise training labels) can be seen as instances of the proposed framework with $\rv{B}=\rv{Z}$ and $\tilde{T}=I\otimes T_{\rv{Y}|\tilde{\rv{Y}}}\in\tri(\tilde{\set{Z}},\set{Z})$. Additionally, note that the proposed framework can encompass combinations of the schemes described above. For instance, supervision schemes in which features at test and labels at training are less precise than those at training and test, respectively, can be seen as instances of the proposed framework with $\rv{B}=(\rv{X},\tilde{\rv{Y}})$, $T=I\otimes T_{\tilde{\rv{Y}}|\rv{Y}}\in\tri(\set{Z},\set{B})$, and $\tilde{T}=T_{\rv{X}|\tilde{\rv{X}}}\otimes I\in \tri(\tilde{\set{Z}},\set{B})$. \begin{figure*} 
	\centering
%	\psfrag{Random variable values}[c][][1]{\hspace{-1mm}RV values}
\psfrag{a}[l][][1]{\hspace{-3.5mm}$\tri(\set{Z})$}
\psfrag{b1}[l][][1]{\hspace{-3.5mm}$\tri(\tilde{\set{Z}}_1)$}
\psfrag{b2}[l][][1]{\hspace{-3.5mm}$\tri(\tilde{\set{Z}}_2)$}
\psfrag{c1}[l][][1]{\hspace{-3.5mm}$\tri(\set{B}_1)$}
\psfrag{c2}[l][][1]{\hspace{-3.5mm}$\tri(\set{B}_2)$}
\psfrag{t1}[l][][1]{\hspace{0mm}$T_1$}
\psfrag{t2}[l][][1]{\hspace{0mm}$T_2$}
\psfrag{s1}[l][][1]{\hspace{0mm}$\tilde{T}_1$}
\psfrag{s2}[l][][1]{\hspace{0mm}$\tilde{T}_2$}
\psfrag{p}[l][][0.8]{\hspace{-1.65mm}$P$}
\psfrag{qa}[l][][0.8]{\hspace{-1.5mm}$\tilde{P}_1$}
\psfrag{qb}[l][][0.8]{\hspace{-1.5mm}$\tilde{P}_2$}
\psfrag{u1}[l][][0.8]{\hspace{-1.5mm}$\set{U}$}
\psfrag{u2}[l][][0.8]{\hspace{-2mm}$\tilde{\set{U}}_1$}
\psfrag{u3}[l][][0.8]{\hspace{-2mm}$\tilde{\set{U}}_2$}
\includegraphics[width=0.95\textwidth]{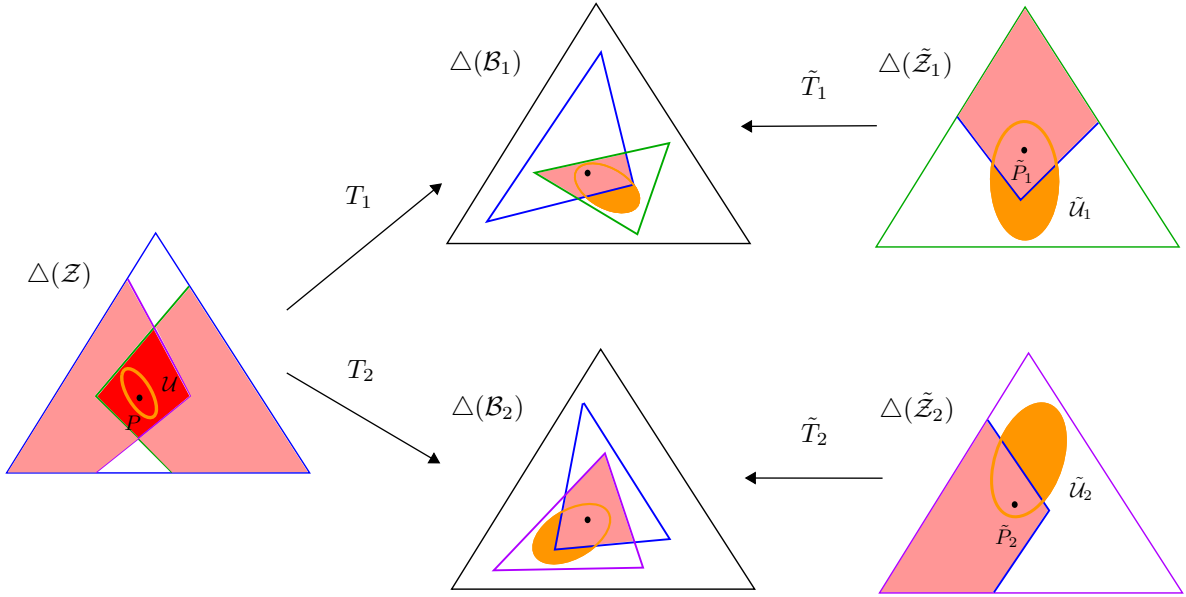}\label{fig2}
	%%\resizebox{0.49 xtwidth}{!}{\includegraphics[draft=false]{Figures/test.eps}}
		\caption{The relationships between test and each type of training variables impose structural constraints for feasible distributions (red polygons), and enable to use information from heterogeneous training data as uncertainty sets (orange ellipses).\label{fig2}}
\end{figure*}

Other current techniques such as those developed under the paradigm of ``covariate shift'' also exploit a specific relationship between the probability distributions of examples at test and training \cite{SugKaw:12}. Those techniques assume that variables at test and training share the same conditional distribution of labels given features, but features at test and training have different marginal distributions $P(x)$ and $\tilde{P}(x)$, i.e., $\set{X}=\tilde{\set{X}}$,  $\set{Y}=\tilde{\set{Y}}$, and
\begin{align}\label{shift} P(x,y)=\tilde{P}(x,y)\frac{P(x)}{\tilde{P}(x)}\,.\end{align}
Such techniques use samples of features at test and training to estimate the function $P(x)/\tilde{P}(x)$, and determine the classification rule using a \ac{ERM} that weights training samples according to the estimated function. Note that \eqref{shift} can be thought of as a mapping of $\tilde{P}$ to $P$. However, such mapping depends on the specific probability distributions followed by test and training features so its usage requires to estimate such mapping for each specific probability distributions. 

\section{Supervision with heterogeneous training data}\label{Sec_gen_het}
In this section we consider supervision cases in which training data is an ensemble of samples with $m$ different types, that is, $d=(d_1,d_2,\ldots,d_m)$, and, for $i=1,2,\ldots,m$, $d_i=(\tilde{z}_i^{(1)},\tilde{z}_i^{(2)},\ldots,\tilde{z}_i^{(n_i)})$ where $\tilde{z}_i^{(j)}$ for $j=1,2,\ldots,n_i$ are independent instantiations of $\tilde{\rv{Z}}_i$. 
% that use different types of training data samples. Specifically, there are $m$ \acp{RV} $\tilde{\rv{X}}_1,\tilde{\rv{X}}_2,\ldots, \tilde{\rv{X}}_m$ describing $m$ different types of training samples, 
Several current supervision schemes use the following ensembles of training data:
\begin{itemize}
\itemsep0em
\item Semi-supervised classification \cite{ChaZie:05,QiTiaShi:12}: a subset of training examples miss labels.%\item supervised domain adaptation (transfer learning) [survey]: some training examples belong to a different domain.
\item \ac{TRS} missing features \cite{ShiBhaSmo:06}: some training examples miss different features' components.
\item Variable quality data \cite{CraKeaWor:06,RooWill:18}: different subsets of training examples are affected by different noise intensities.
\item Domain adaptation with multiple sources \cite{ManMohRos:09}: different subsets of training examples belong to different but similar domains.
\end{itemize}
The following shows how heterogenous training data can be aggregated by further extending the framework presented in previous section.
%supervised classification can be carried out with heterogeneous training data by further extending the framework described above. 
\begin{table*}
\caption{Current heterogeneous supervision schemes}\label{table-het}
\centering
\begin{tabular}{llll}
\hline\hline
Supervision scheme					&\hspace{-0.2cm}\begin{tabular}{l}Training data\\ types $\tilde{\rv{Z}}_i$\end{tabular}			&Bridges $\rv{B}_i$					&\hspace{-0.2cm}\begin{tabular}{l}Prob.\\ transformations\end{tabular}\\\hline\midrule
Semi-supervision&\hspace{-0.2cm}\begin{tabular}{l}$\tilde{\rv{Z}}_1=\rv{Z}=(\rv{X},\rv{Y})$\\$\tilde{\rv{Z}}_2=\rv{X}$\end{tabular}&\hspace{-0.2cm}\begin{tabular}{l}$\rv{B}_1=\rv{Z}=\tilde{\rv{Z}}_1$\\$\rv{B}_2=\tilde{\rv{Z}}_2$\end{tabular}\hspace{-0.2cm}&\hspace{-0.2cm}\begin{tabular}{l}$T_1=\tilde{T}_1=I$\hspace{-0.2cm}\\$T_2=T_{\rv{X}|\rv{Z}}$,\  $\tilde{T}_2=I$\end{tabular}\\\midrule
\ac{TRS} missing features	&\hspace{-0.2cm}\begin{tabular}{l}$\tilde{\rv{Z}}_1=\rv{Z}=(\rv{X},\rv{Y})$\\$\tilde{\rv{Z}}_{i+1}=(\bar{\rv{X}}_i,\rv{Y})$\end{tabular}	&\hspace{-0.2cm}\begin{tabular}{l}$\rv{B}_1=\rv{Z}=\tilde{\rv{Z}}_1$\\$\rv{B}_{i+1}=\tilde{\rv{Z}}_{i+1}$\end{tabular}\hspace{-0.2cm}&\hspace{-0.2cm}\begin{tabular}{l}$T_1=\tilde{T}_1=I$\\$T_{i+1}=T_{\bar{\rv{X}}_i|\rv{X}}\otimes I$\\ $\tilde{T}_{i+1}=I$		\end{tabular}\\\midrule
Variable quality data 	&\hspace{-0.2cm}\begin{tabular}{l}$\tilde{Z}_i=(\rv{X},\rv{Y}_i)$, $y_i$ noisy\\ $\rv{Y}_i\neq\rv{Y}_j$, $i\neq j$ 	\end{tabular}\hspace{-0.2cm}								& 	$\rv{B}_i=\tilde{\rv{X}}_i$		&\hspace{-0.2cm}\begin{tabular}{l}$T_i=I\otimes T_{\tilde{\rv{Y}}_i|\rv{Y}}$\\$\tilde{T}_i=I$\end{tabular}										\\\midrule
\hspace{-0.2cm}\begin{tabular}{l}Domain adaptation with\\multiple sources \end{tabular} &$\tilde{\set{X}}_i=\set{X}$, $\tilde{\rv{X}}_i\neq\rv{X}$ &General&$T_i=\tilde{T}_i=T_{\rv{B}|\rv{Z}}$\\\hline\hline
\end{tabular}
\end{table*}
Let, for $i=1,2\ldots,m$, $\set{B}_i$ be a support set, and $T_i\in\tri(\set{Z},\set{B}_i)$ and $\tilde{T}_i\in\tri(\tilde{\set{Z}}_i,\set{B}_i)$ be probabilistic transformations such that $T_i(P)=\tilde{T}_i(\tilde{P}_i)$ for $P$ and $\tilde{P}_i$ the distributions of $\rv{Z}$ and $\tilde{\rv{Z}}_i$, respectively. Analogously to the case described in previous section for only one type of training data, i.e., $m=1$, such probabilistic transformations allow to extract the information in heterogeneous and non-standard training data (see Fig.~\ref{fig2}). Specifically, feasible distributions and uncertainty sets in $\tri(\set{Z})$ can be defined as
\begin{align*}\set{F}=\{&Q\in\tri(\set{Z}):\  \exists\, \tilde{Q}_i\in\tri(\tilde{Z}_i),\\&   T_i(Q)=\tilde{T}_i(\tilde{Q}_i),\  i=1,2,\ldots,m\}\\
\set{U}=\{&Q\in\tri(\set{Z}):\  \sum_{i=1}^mw_i\psi(T_i(Q),\tilde{T}_i(\tilde{P}_{e_i}))<\varepsilon\}
\end{align*}
with $w_i>0$ a parameter weighting the discrepancy in each $\tri(\set{B}_i)$, e.g., $w_i\propto \sqrt{n}_i$. Therefore, learning from non-standard heterogeneous training data $d_1,d_2,\ldots,d_m$ can be approached by \ac{GRRM} generalizing equation \eqref{gen_supervision} as 
\begin{align}\label{het_supervision}\min_{Q\in\set{F}}\sum_{i=1}^m w_i\psi\big(T_i(Q),\tilde{T}_i(\tilde{P}_{e_i})\big)-\lambda H(Q)\end{align}
where $\lambda>0$ is a regularization parameter. %We next describe how the above framework covers existing supervision schemes and can give rise to novel supervision schemes of practical relevance.

% quantifying the rate of change of the entropy in terms of the uncertainty set sizes .
%xpected disprepancy between $\tilde{T}_i(\tilde{p}_{d_i})$ and the actual probability distribution of $\rv{B}$ (see Fig.~\ref{fig2}).

%\subsection{Different heterogeneous supervision schemes under the proposed framework}
Table~\ref{table-het} shows how different current supervision schemes with heterogeneous training data can be addressed under the proposed framework. In semi-supervision and \ac{TRS} missing features, samples in one subset of the training data follow the same distribution as those at test stage, i.e., $\rv{B}_1=\tilde{\rv{Z}}_1=\rv{Z}$, while the remaining training samples are less precise than those at test, i.e.,  $\rv{B}_i=\tilde{\rv{Z}}_i$ and $T_i=T_{\tilde{\rv{Z}}_i|\rv{Z}}\in\tri(\set{Z},\tilde{\set{Z}}_i)$ for $i>1$. In particular, for \ac{TRS} missing features, training data can be classified in terms of the feature component that is missing with $\bar{x}_i=(x_1,x_2,\ldots,x_{i-1},x_{i+1},\ldots,x_r)$. In other supervision schemes, such as variable quality data or domain adaptation with multiple sources, the training data subsets are affected by different label noises ($T_i=I\otimes T_{\tilde{\rv{Y}}_i|\rv{Y}}$) or belong to different domains with a common representation ($T_i=\tilde{T}_i=T_{\rv{B}|\rv{Z}_i}$), respectively.

The proposed framework can also enable the development of novel supervision schemes that aggregate general ensembles of training data, such as those described in fourth case study in Section~\ref{Sec_exp}. These new supervision schemes could be specially suitable for environments of open collaboration where each participant in the annotation process could choose a type of contribution based on resources, commitment, remuneration, etc. For instance, different groups of participants could choose to use high- or low-resolution features, to annotate examples quickly or meticulously, etc.
\section{Experiments}\label{Sec_exp}
This section shows the feasibility of the general framework proposed to encompass multiple existing schemes as well as to enable novel types of supervision. Specifically, we consider four experimentation case studies: two well-studied non-standard supervision schemes, and two novel non-standard supervision schemes.
In particular, we solved the convex optimization problems \eqref{gen_supervision} and \eqref{het_supervision} using CVX package \cite{GraBoyYe:06} with 0-1 entropy given by \eqref{01ent}. As in \cite{FarTse:16}, the distributions considered have features support that coincides with that of the empirical distribution, and we use the discrepancy given by \eqref{tse} (more experimentation details can be found in the Appendix of the supplementary material).

Table~\ref{table3} shows the estimated accuracy of proposed \ac{GRRM} for two existing supervision schemes (noisy labels and semi-supervision) in comparison with several representative methods using 3 UCI datasets.  In this two case studies we used \eqref{tse} with statistic $t(z)=(\theta_-(y),\theta_-(y)x,\theta_+(y),\theta_+(y)x)$, where $\theta=(\theta_-,\theta_+)$ is the one-hot encoding \cite{FarTse:16} of the class $y$ and the step \eqref{min_risk} is solved by a \ac{SVM} with weights given by the solutions of \eqref{gen_supervision} and \eqref{het_supervision}. For noisy labels we compare the accuracy of \ac{GRRM} with that of 4 methods, as reported in \cite{LiuDac:16} (case $\rho^-=0.1$ and $\rho^+=0.3$). For semi-supervision we compare the accuracy of \ac{GRRM} with that of 3 methods, as reported in \cite{QiTiaShi:12}, as well as method SMIR\footnote{Implemented using code in https://github.com/wittawatj/smir} proposed in \cite{NiuJitDai:13} (we used 5\% and 30\% labeled and unlabeled samples, resp.). The results in Table~\ref{table3} show that \ac{GRRM} can obtain state-of-the-art accuracies in well-studied non-standard supervision schemes.

\begin{table}%[h]
\caption{Accuracy of proposed \ac{GRRM} for existing supervision schemes.\label{table3}}
\centering
\begin{tabular}{lllll}
\hline\hline
&\multirow{2}{*}{Technique}&\multicolumn{3}{c}{Data set}\\
&&German&Heart&Diabetes\\
\multirow{7}{*}{\rotatebox{90}{Noisy labels}}&&&&\\&\ac{GRRM}&72.6\%&78.3\%&73.2\%\\
&IW&69.6\%&72.1\%&71.5\%\\
&LD&70.8\%&72.2\%&73.2\%\\
&eIW&68.8\%&70.1\%&74.3\%\\
&StPMKL&67.2\%&54.7\%&66.5\%\\&&&&\\\hline\\
\multirow{7}{*}{\rotatebox{90}{Semi-supervision}}&&&&\\&\ac{GRRM}&70.0\%&77.8\%&70.0\%\\
&Lap-TSVM&63.5\%&75.8\%&63.4\%\\
&Lap-SVM&64.6\%&74.3\%&63.0\%\\
&TSVM&61.2\%&73.7\%&60.0\%\\
&SMIR&70.0\%&75.1\%&68.6\%\\&&&&\\\hline\hline
\end{tabular}
\end{table}
\begin{figure}%[H]
%\psfrag{73}[][][0.45]{\hspace{-1mm}73}
\psfrag{Y}[l][t][0.7]{\hspace{-7mm}Accuracy}
\psfrag{X}[l][b][0.7]{\hspace{-20mm}Noise probabilities $\rho^+$ and $\eta$}
\psfrag{D123456789012}[l][][0.5]{\hspace{-12.4mm}Benchmark bound}
\psfrag{D3}[l][][0.5]{\hspace{-0.9mm}Naive}
\psfrag{D4}[l][][0.5]{\hspace{-1mm}\ac{GRRM}}
\subfigure[Supervision with noisy labels (training) and noisy features (test).\label{fig_noisy}]{
\includegraphics[width=0.47\textwidth]{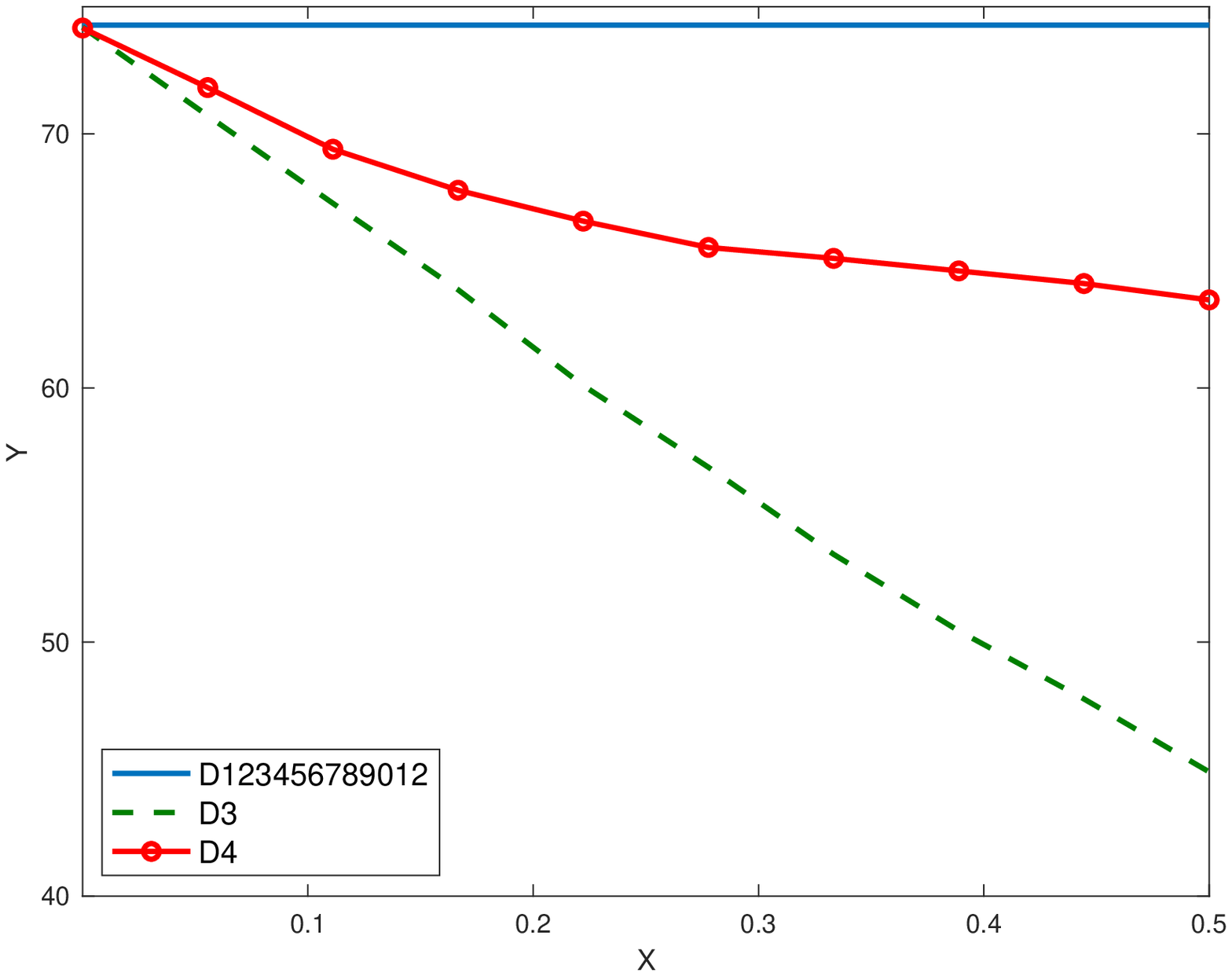}}
\hfill
\psfrag{Y}[l][t][0.7]{\hspace{-7mm}Accuracy}
\psfrag{X}[l][b][0.7]{\hspace{-19mm}Additional training samples}
\psfrag{D123456789012}[l][][0.5]{\hspace{-12.4mm}Standard}
\psfrag{D3}[l][][0.5]{\hspace{-1mm}Noisy labels}
\psfrag{D4}[l][][0.5]{\hspace{-1mm}Dom. adaptation}
\psfrag{D5}[l][][0.5]{\hspace{-1mm}Priv. information}
\subfigure[Learning curves using training data with 4 different types.\label{fig_het}]
{\includegraphics[width=0.47\textwidth]{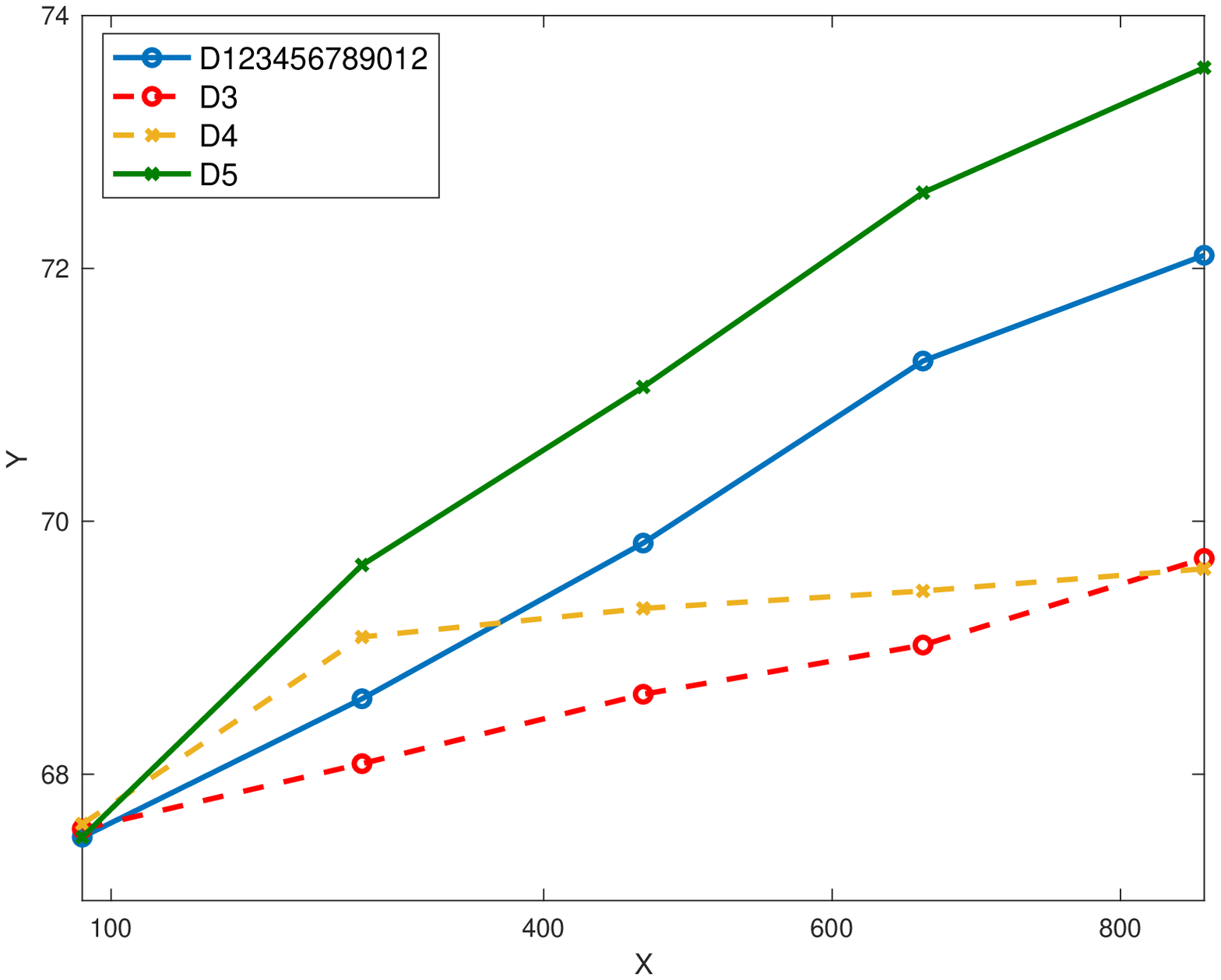}}
\end{figure}

%\begin{table}%[h]
%\centering
 % \caption{Data sets}
 % \label{table:data}
%  \begin{tabular}{lll}
%  \toprule
 %   Name   			& num. features	& num. instances\\
%  \midrule
 %   Breast cancer 	& 9 			& 263 (77+,186-)\\
  %  Diabetes 		& 8 			& 768 (268+,500-)\\
  %  German 			& 20 			& 1000 (300+,700-)\\
  %  Heart 			& 13 			& 270 (120+,150-)\\
 % Tic-tac-toe		& 9			& 958 (626+,332-)\\
 % \bottomrule
%  \end{tabular}
%\end{table}

%Figure~\ref{fig_priv} shows the accuracy of the proposed \ac{GRRM} for supervision with privileged information in comparison with a \ac{SVM} that does not use privileged information, for the time series used in \cite{PecVap:10} where \ac{GRRM}-2 and \ac{GRRM}-4 use 2 and 4 future observations. Figure~\ref{fig_semi} shows the accuracy of the proposed \ac{GRRM} for semi-supervision in comparison with a \ac{SVM} that does not use unlabeled training data and the semi-supervision method proposed in \cite{QiTiaShi:12}  (Lap-TSVM) in the ``Heart'' UCI dataset, for different ratios between training samples without and with labels. 

Fig.~\ref{fig_noisy} and Fig.~\ref{fig_het} show the accuracy of proposed \ac{GRRM} in novel non-standard supervision schemes using the UCI tic-tac-toe dataset. In particular, the board configurations in the 2x2 upper-left block are used as features to predict the game end, and classification is done by computing labels' conditional probabilities.

The first novel supervision scheme considers noisy labels at training and noisy features at test. We compare classification accuracy with varying probabilities of errors for 3 implementations: benchmark bound obtained by using \ac{ERM} with noiseless features and labels, naive \ac{ERM} that does not account for the noises, and proposed \ac{GRRM} using \eqref{tse} with indicator functions of each board case as statistics. The probabilities of incorrectly labeling a ``win for x'' and a ``not win for x'' are $\rho^+$ and $\rho^-$, respectively, while the probability of an error in reading each board's cell is $\eta$. Fig.~\ref{fig_noisy} compares the accuracies obtained varying $\rho^+$ and $\eta$ from $0$ to $0.5$ with $\rho^+=\eta$ and $\rho^-=\rho^+/2$. It can be observed that proposed \ac{GRRM} can enable the usage of both noisy labels at training and noisy features at test even when they are severely affected by noise.

The second novel supervision scheme aggregates training data with 4 different types: standard supervision, noisy labels ($\rho^-=0.1$, $\rho^+=0.3$), domain adaptation with the middle vertical 3x1 block as features, and privileged information with all cells except the up-right and low-left corners as features. Fig.~\ref{fig_het} compares the accuracies obtained by proposed \ac{GRRM} using different amounts of training data for each type. The leftmost points in the curves show the accuracy obtained aggregating $80$ samples of each type, and the remaining points show how accuracy increases by increasing the number of training samples of different types while keeping the others fixed. It can be observed that the proposed \ac{GRRM} can aggregate training data with different types. As expected, the accuracy increases faster by adding more informative training samples (standard and privileged information) than by adding less informative training samples (noisy labels and domain adaptation). These heterogenous supervision schemes can improve the accuracy vs cost trade-off in training stages by enabling the aggregation of multiple samples' types with different acquisition costs and information contents. 

%\begin{figure}%[H]
%\psfrag{73}[][][0.45]{\hspace{-1mm}73}

%\caption{Learning curves using training data with 4 different types.\label{fig_het}}
%\end{figure}
\section{Conclusion}
The paper presents a unifying framework and learning techniques for supervised classification with non-standard and heterogenous training data. The introduced methodology of generalized robust risk minimization (GRRM) can enable to develop learning algorithms for current and novel supervision schemes in a unified manner. The results presented can lead to new learning scenarios able to balance cost vs accuracy trade-offs of training stages, and seamlessly aggregate ensembles of training data with assorted types and qualities. %Finally, the presented \ac{GRRM} also puts forward new research challenges such as the generalization of efficient optimization strategies as those developed for \ac{RRM}, specially in regimes with high-dimensional or massive data. 

%\resizebox{0.49\columnwidth}{!}{
%\begin{minipage}{0.49\linewidth}
%\vspace{0pt}
%\begin{figure}
%\caption{Accuracy means and standard\\ deviations for noisy labels supervision}\label{table-noisy}
%\begin{minipage}{0.49\linewidth}
%\vspace{0pt}
%\begin{tabular}{l c c c}\hline\hline
%Dataset&LD&IW&\ac{GRRM}\\\hline\hline
%Breast&70.6$\pm$5.2&72.7$\pm$5.1&71.7$\pm$2.6\\
%Diabetis&73.2$\pm$2.4&71.5$\pm$3.0&73.2$\pm$5.4\\
%German&70.8$\pm$2.9&69.6$\pm$2.4&72.6$\pm$3.9\\
%Heart&72.2$\pm$8.2&72.1$\pm$6.7&78.3$\pm$6.9\\\hline\hline
%\end{tabular}
%\end{table}

%\end{minipage}
%\begin{minipage}{0.55\linewidth}
%\vspace{0pt}
%\begin{figure}
	%\centering
%	\psfrag{Random variable values}[c][][1]{\hspace{-1mm}RV values}

%\section{Conclusion}
%The unifying framework presented in the paper can approach different supervision schemes in a unified manner 

%\small

%\bibliographystyle{unsrtnat}
%\bibliographystyle{IEEEtran}
\newpage
%\bibliography{manuscript}

%\bibliographystyle{unsrt}

\end{document}